\def\year{2022}\relax
\documentclass[letterpaper]{article} 
\usepackage{aaai22}  
\usepackage{times}  
\usepackage{helvet}  
\usepackage{courier}  
\usepackage[hyphens]{url}  
\usepackage{graphicx} 
\usepackage{natbib}  
\usepackage{caption} 
\DeclareCaptionStyle{ruled}{labelfont=normalfont,labelsep=colon,strut=off} 
\usepackage{algorithm}
\usepackage{algorithmic}
\usepackage{multirow}
\usepackage{subfigure}
\usepackage{makecell}
\usepackage{color}
\usepackage{amsmath}
\usepackage{amssymb}
\usepackage{gensymb}

\usepackage{newfloat}
\usepackage{listings}
\floatstyle{ruled}
\newfloat{listing}{tb}{lst}{}
\floatname{listing}{Listing}
\title{MetaNODE: Prototype Optimization as a Neural ODE for Few-Shot Learning}
\author{
    Baoquan Zhang,
    Xutao Li$^*$,
    Shanshan Feng,
    Yunming Ye\thanks{Corresponding author.},
    Rui Ye
}
\affiliations{
	Harbin Institute of Technology, Shenzhen \\
	zhangbaoquan@stu.hit.edu.cn,
	\{lixutao, victor\_fengss, yeyunming\}@hit.edu.cn,
	yerui\_hitsz@163.com

}

\usepackage{bibentry}

\begin{document}

\maketitle

\begin{abstract}
Few-Shot Learning (FSL) is a challenging task, \emph{i.e.}, how to recognize novel classes with few examples? Pre-training based methods effectively tackle the problem by pre-training a feature extractor and then predicting novel classes via a cosine nearest neighbor classifier with mean-based prototypes. Nevertheless, due to the data scarcity, the mean-based prototypes are usually biased. In this paper, we attempt to diminish the prototype bias by regarding it as a prototype optimization problem. To this end, we propose a novel meta-learning based prototype optimization framework to rectify prototypes, \emph{i.e.}, introducing a meta-optimizer to optimize prototypes. Although the existing meta-optimizers can also be adapted to our framework, they all overlook a crucial gradient bias issue, \emph{i.e.}, the mean-based gradient estimation is also biased on sparse data. To address the issue, we regard the gradient and its flow as meta-knowledge and then propose a novel Neural Ordinary Differential Equation (ODE)-based meta-optimizer to polish prototypes, called MetaNODE. In this meta-optimizer, we first view the mean-based prototypes as initial prototypes, and then model the process of prototype optimization as continuous-time dynamics specified by a Neural ODE. A gradient flow inference network is carefully designed to learn to estimate the continuous gradient flow for prototype dynamics. Finally, the optimal prototypes can be obtained by solving the Neural ODE. Extensive experiments on miniImagenet, tieredImagenet, and CUB-200-2011 show the effectiveness of our method. Our code is available \footnote {\url{https://github.com/zhangbq-research/MetaNODE}}. 
\end{abstract}

\section{Introduction}
With abundant annotated data, deep learning techniques have shown very promising performance for many applications, \emph{e.g.}, image classification \cite{he16}. However, preparing enough annotated samples is very time-consuming, laborious, or even impractical in some scenarios, \emph{e.g.}, cold-start recommendation \cite{zheng2021} and medical diagnose \cite{PrabhuKRCSA19}. Few-shot learning (FSL), which aims to address the issue by mimicking the flexible adaptation ability of human to novel tasks from very few examples, has been proposed and received considerable attentions. Its main rationale is to learn meta-knowledge from base classes with sufficient labeled samples and then employ the meta-knowledge to perform class prediction for novel classes with scarce examples \cite{LiLX0019}.

\begin{figure}
	\centering
	\subfigure[Prior Works]{ 
		\label{fig0a} 
		\includegraphics[width=0.43\columnwidth]{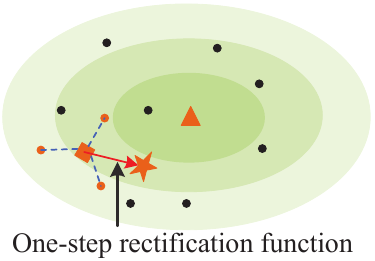}}
	\subfigure[Our Method]{ 
		\label{fig0b} 
		\includegraphics[width=0.43\columnwidth]{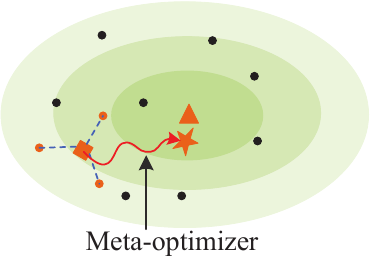}}
	\caption{Pre-training based method estimates prototypes in an average manner, which suffers from a prototype bias issue. Prior works diminish the bias in a one-step manner (a). Our method addresses it in a meta-optimization manner (b). Here, orange and black points denote training and test samples, respectively. Orange square, star, and triangle denotes the mean-based, rectified, and real prototypes, respectively. }
	\vspace{-9.5pt}
	\label{fig0}
\end{figure}

Previous studies primarily address the FSL problem using the idea of meta-learning, \emph{i.e.}, constructing a large set of few-shot tasks on base classes to learn task agnostic meta-knowledge \cite{FlennerhagRPVYH20}. Recently, Chen et al. regard feature representation as meta-knowledge, and propose a simple pre-training method \cite{Yinbo20}, which delivers more promising performance. In the method, they first pre-train a feature extractor on all base classes, and then perform novel class prediction via mean-based prototypes. However, this method suffers from a {\bf prototype bias issue}, \emph{i.e.}, the discrepancy between calculated mean and real prototypes. As illustrated in Figure~\ref{fig0}, the mean-based prototype (orange square) is usually far away from real prototype (triangle). This kind of prototype bias is caused by the fact that scarce labeled samples cannot provide a reliable mean estimation for the prototypes \cite{YaohuiWang_pr}. To address the drawback, some prior works attempt to learn a one-step prototype rectification function from a large set of few-shot tasks \cite{YaohuiWang_pr, XueW20, zhang2021}, which is shown in Figure~\ref{fig0a}. However, characterizing the prototype bias with a one-step rectification function is too coarse to obtain accurate prototypes (as we will see in Table~\ref{table3} of the Section~\ref{section_4_4}).

In this paper, we propose a novel meta-learning based prototype optimization framework to rectify the prototype bias. In the framework, instead of using the one-step rectification manner, we consider the bias reduction as a prototype optimization problem and attempt to diminish the prototype bias with an optimization-based meta-learning method (called meta-optimizer). The idea behind such design is learning a novel Gradient Descent Algorithm (GDA) on base classes and then utlizing it to polish the prototypes via a few gradient update steps for novel classes. Specifically, we first pre-train a classifier on all base classes to obtain a good feature extractor. Then, given a few-shot task, as shown in Figure~\ref{fig0b}, we average the extracted features of all labeled samples as the initial prototype for each class. As a sequel, these prototypes will be further optimized to reduce the prototype bias through a meta-optimizer. Finally, we perform the class prediction via a nearest neighbor classifier. 

The workhorse of our framework is the meta-optimizer, in which the mean-based prototypes will be further polished through GDA. Even though the existing meta-optimizer such as ALFA \cite{BaikCCKL20} and MetaLSTM \cite{RaviL17} can also be utilized for this purpose, 
they all suffer from a common drawback, called {\bf gradient bias issue}, \emph{i.e.}, their gradient estimation is inaccurate on sparse data. The issue appears because all the existing meta-optimizers carefully model the hyperparameters (\emph{e.g.}, initialization \cite{RaghuRBV20} and regularization parameters \cite{BaikCCKL20, FlennerhagRPVYH20}) in GDA as meta-knowledge, but roughly estimate the gradient in an average manner with very few labeled samples, which is usually inaccurate. Given that the gradient estimation is inaccurate, more excellent hyperparameters are not meaningful and cannot lead to a stable and reliable prototype optimization. 

To address the issue, we treat the gradient and its flow in GDA as meta-knowledge, and propose a novel Neural Ordinary Differential Equation (ODE)-based meta-optimizer to model the process of prototype optimization as continuous-time dynamics specified by a Neural ODE, called MetaNODE. The idea is inspired by the fact that the GDA formula is indeed an Euler-based discrete instantiation of an ODE \cite{bu2020}, and the ODE will turn into a Neural ODE \cite{ChenRBD18} when we treat its gradient flow as meta-knowledge. The advantage of such meta-optimizer is the process of prototype rectification can be characterized in a continuous manner, thereby producing more accurate prototypes for FSL.
Specifically, in the meta-optimizer, a gradient flow inference network is carefully designed, which learns to estimate the continuous-time gradient flow for prototype dynamics. Then, given an initial prototype (\emph{i.e.}, the mean-based prototype), the optimal prototype can be obtained by solving the Neural ODE for FSL. 

Our main contributions can be summarized as follows:

\begin{itemize}
	\item We propose a new perspective to rectify prototypes for FSL, by regarding the bias reduction problem as a prototype optimization problem, and present a novel meta-learning based prototype optimization framework. 
	\item We identify a crucial issue of existing meta-optimizers
	, \emph{i.e.}, gradient bias issue, which impedes their applicability to our framework. To address the issue, we propose a novel Neural ODE-based meta-optimizer (MetaNODE) by modeling the process of prototype optimization as continuous-time dynamics specified by a Neural ODE. Our MetaNODE optimizer can effectively alleviate the gradient bias issue and leads to more accurate prototypes. 
	\item We conduct comprehensive experiments on both transductive and inductive FSL settings, which demonstrate the effectiveness of our method.

\end{itemize}

\section{Related Work}

\subsection{Few-Shot Learning}
FSL is a challenging task, aiming to recognize novel classes with few labeled samples. According to the test setting, FSL can be divided into two groups, \emph{i.e.}, inductive FSL and transductive FSL. The former assumes that information from test data cannot be utilized when classifying the novel class samples while the latter considers that all the test data can be accessed to make novel class prediction.

In earlier studies, most methods mainly focus on inductive FSL setting, which can be roughly grouped into three categories.  1) Metric-based approaches. This line of works focuses on learning a task-agnostic metric space and then predicting novel classes by a nearest-centroid classifier with Euclidean or cosine distance such as \cite{NguyenLWKRJ20, SnellSZ17, li2019finding}. 
2) Optimization-based approaches. The key idea is to model an optimization algorithm (\emph{i.e.}, GDA) over few labeled samples within a meta-learning framework \cite{Johannes21, RaghuRBV20}, which is known as meta-optimizer, such as MetaLSTM \cite{RaviL17} and ALFA \cite{BaikCCKL20}. 3) Pre-training based approaches. This type of works mainly utilizes a two-phases training manner to quickly adapt to novel tasks, \emph{i.e.}, pre-training and fine-tuning phases, such as \cite{LiuCLL0LH20, ShenLQSC21, Yinbo20, Zhiuncertainty21, LiFG21}. 

Recently, several studies explored transductive FSL setting and showed superior performance on FSL, which can be divided into two categories. 1) Graph-based approaches. This group of methods attempts to propagate the labels from labeled samples to unlabeled samples by constructing an instance graph for each FSL task, such as \cite{Chen_2021_CVPR, YangLZZZL20, Tang_2021_CVPR}.  
2) Pre-training based approaches. This kind of studies still focuses on the two-stages training paradigms. Different from inductive FSL methods, these methods further explore the unlabeled samples to train a better classifier \cite{Malik20, HuMXSOLD20, WangXLZF20, ZikoDGA20} or construct more reliable prototypes \cite{XueW20, zhang2021}. 
In this paper, we also target at obtaining reliable prototypes for FSL. Different from these existing methods, we regard it as a prototype optimization problem and propose a new Neural ODE-based meta-optimizer for optimizing prototypes. 
\begin{figure*}[t]
	\centering
	\includegraphics[width=1.0\textwidth]{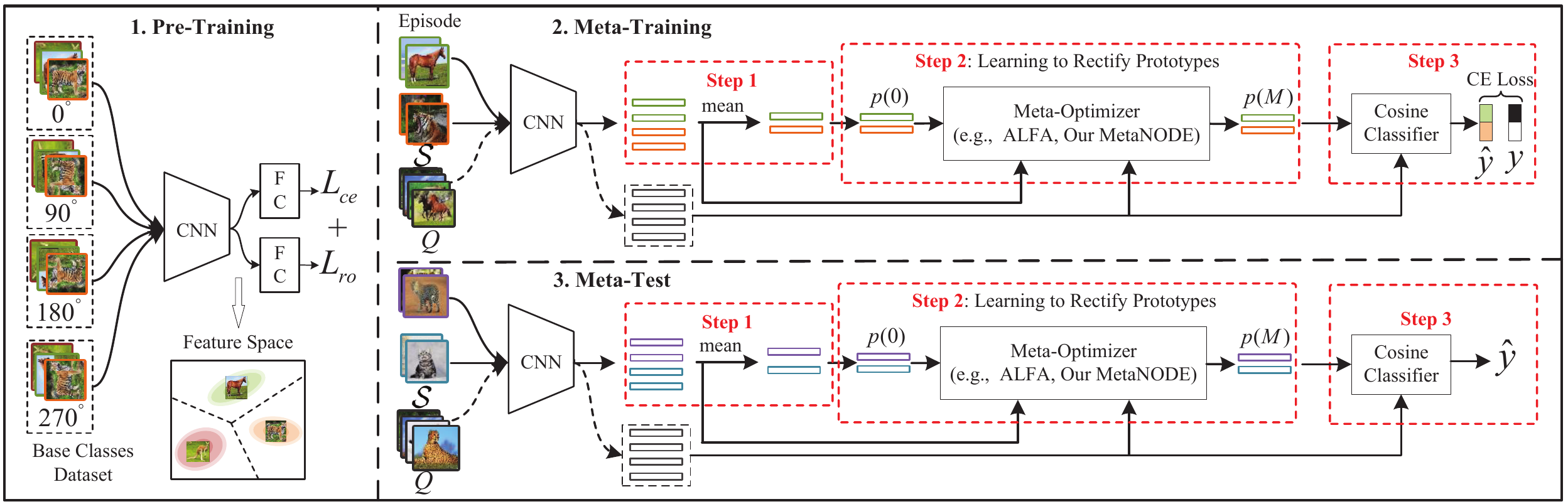} 
	\caption{The meta-learning based prototype optimization framework, which rectifies prototypes in a meta-optimization manner.}
	\vspace{-12.5pt}
	\label{fig2}
\end{figure*}

\subsection{Neural ODE}
Neural ODE, proposed by \cite{ChenRBD18}, is a continuous-time model, aiming to capture the evolution process of a state by representing its gradient flow with a neural network. Recently, it has been successfully used in various domains, such as irregular time series prediction \cite{RubanovaCD19}, knowledge graph forecasting \cite{abs210105151}, MRI image reconstruction \cite{ChenCS20}, and image dehazing \cite{ShenL0X020}. However, to our best knowledge, there is few previous works to explore it for FSL. In this paper, we propose a Neural ODE-based meta-optimizer to polish prototypes. Its advantage is that prototype dynamics can be captured in a continuous manner, which produces more accurate prototypes for FSL. 

\section{Methodology}



\subsection{Problem Definition}
\label{section_3_1}
For a $N$-way $K$-shot problem, two datasets are given: a base class dataset $\mathcal{D}_{base}$ and a novel class dataset $\mathcal{D}_{novel}$. The base class dataset $\mathcal{D}_{base}=\{(x_i, y_i)\}_{i=0}^{B}$ is made up of abundant labeled samples, where each sample $x_i$ is labeled with a base class $y_i \in \mathcal{C}_{base}$ ($\mathcal{C}_{base}$ denotes the set of base classes). The novel class dataset consists of two subsets: a training set $\mathcal{S}$ with few labeled samples (called support set) and a test set $\mathcal{Q}$ consisting of unlabeled samples (called query set). Here, the support set $\mathcal{S}$ is composed of $N$ classes sampled from the set of novel class $\mathcal{C}_{novel}$, and each class only contains $K$ labeled samples. Note that the base class set and novel class set are disjoint, \emph{i.e.}, $\mathcal{C}_{base} \cap \mathcal{C}_{novel} = \emptyset$. 

For transductive FSL, we regard all query samples $x \in \mathcal{Q}$ as unlabeled sample set $\mathcal{Q}'$. Our goal is to learn a classifier for query set $\mathcal{Q}$ by leveraging unlabeled sample set $\mathcal{Q}'$, support set $\mathcal{S}$, and base class dataset $\mathcal{D}_{base}$. However, for inductive FSL, the classifier is obtained only by leveraging $\mathcal{S}$ and  $\mathcal{D}_{base}$. In the following subsections, we mainly focus on transductive FSL to introduce our method, and how to adapt to inductive FSL will be explained at the end of the section. 

\subsection{Prototype Optimization Framework}
\label{section3_2}
In this paper, we focus on addressing the prototype bias issue appearing in the pre-training FSL method \cite{Yinbo20}. Different from existing one-step rectification methods \cite{YaohuiWang_pr, XueW20}, we regard the bias reduction as a prototype optimization problem and present a novel meta-learning based prototype optimization framework to rectify prototypes. Our idea is introducing a prototype meta-optimizer to learn to diminish the prototype bias. As shown in Figure~\ref{fig2}, the framework consists of three phases, \emph{i.e.}, pre-training, meta-training, and meta-test phases. Next, we detail on them, respectively.  

\noindent {\bf Pre-Training.} Following \cite{RodriguezLDL20}, we first pretrain a feature extractor $f_{\theta_f}()$ with parameters $\theta_f$ by minimizing both a classification loss $L_{ce}$ and an auxiliary rotation loss $L_{ro}$ on all base classes. This aims to obtain a good image representation. Then, the feature extractor is frozen. 

\noindent {\bf Meta-Training.} 
Upon the feature extractor $f_{\theta_f}()$, we introduce a meta-optimizer $g_{\theta_g}()$ to learn to rectify prototypes in an episodic training paradigm \cite{VinyalsBLKW16}. The idea behind such design is that learning task-agnostic meta-knowledge about prototype rectification from base classes and then applying this meta-knowledge to novel classes to obtain more reliable prototypes for FSL. The main details of the meta-optimizer will be elaborated in Section~\ref{section_3_3}. Here, we first introduce the workflow depicted in Figure~\ref{fig2}.  

As shown in Figure~\ref{fig2}, following the episodic training paradigm \cite{VinyalsBLKW16}, we first mimic the test setting and construct a number of $N$-way $K$-shot tasks (called episodes) from base class dataset $\mathcal{D}_{base}$. For each episode, we randomly sample $N$ classes from base classes $\mathcal{C}_{base}$, $K$ images per class as support set $S$, and $M$ images per class as query set $Q$. Then, we train the  above meta-optimizer $g_{\theta_g}()$ to polish prototypes by minimizing the negative log-likelihood estimation on the query set $\mathcal{Q}$. That is,
\begin{equation}
	\begin{aligned}
		\min\limits_{\theta_g}  \mathbb{E}_{(\mathcal{S},\mathcal{Q}) \in \mathbb{T}} \sum_{(x_i,y_i) \in \mathcal{Q}} -log(P(y_i|x_i, \mathcal{S}, \mathcal{Q}', \theta_g)),
	\end{aligned}
	\label{eq5_0}
\end{equation}
where $\mathbb{T}$ is the set of constructed $N$-way $K$-shot tasks and $\theta_g$ denotes the parameters of the meta-optimizer $g_{\theta_g}()$. Next we introduce how to calculate the class probability $P(y_i|x_i, \mathcal{S}, \mathcal{Q}', \theta_g)$, including the following three steps: 

{\bf Step 1.} We first leverage the feature extractor $f_{\theta_f}()$ to represent each image. Then we compute the mean-based prototype of each class $k$ as initial prototype $p_k(0)$ at $t=0$:
\begin{equation}
	p_k(0) = \frac{1}{|\mathcal{S}_k|} \sum_{(x_i, y_i) \in \mathcal{S}_k} f_{\theta_f}(x_i),
	\label{eq5}
\end{equation}
where $\mathcal{S}_k$ is the support set extracted from class $k$ and $t$ denotes the iteration step (when existing meta-optimizers are employed) or the continous time (when our MetaNODE described in Section~\ref{section_3_3} is utilized). For clarity, we denote the prototype set $\{p_k(t)\}_{k=0}^{N-1}$ as the prototypes $p(t)$ of classifiers at iteration step/time $t$, \emph{i.e.}, $p(t) = \{p_k(t)\}_{k=0}^{N-1}$.

{\bf Step 2:} Unfortunately, the initial prototypes $p(0)$ are biased since only few support samples are available. To eliminate the bias, we view the prototype rectification as an optimization process. Then, given the initial prototypes $p(0)$, the optimal prototypes $p(M)$ can be obtained by leveraging the meta-optimizer $g_{\theta_g}()$ to optimize prototypes. That is, 
\begin{equation}
	p(M) = \Psi(g_{\theta_{g}}(), p(0), \mathcal{S}, \mathcal{Q}', t=M),
	\label{eq6}
\end{equation}where $M$ is the total iteration number/integral time and $\Psi()$ denotes the process of prototype optimization. Please refer to Section~\ref{section_3_3} for the details of the optimization process. 

{\bf Step 3:} Finally, we regard the optimal prototypes $p(M)$ as the final prototypes. Then, we evaluate the class probability that each sample $x_i \in \mathcal{Q}$ belongs to class $k$ by computing the cosine similarity between $x_i$ and $p(M)$. That is,
\begin{equation}
	P(y=k|x_i, \mathcal{S}, \mathcal{Q}', \theta_g) = \frac{e^{\gamma \cdot <f_{\theta_f}(x_i), p_{k}(M)>}}{\sum_c e^{\gamma \cdot <f_{\theta_f}(x_i), p_{c}(M)>}},
	\label{eq7}
\end{equation}
where $<\cdot>$ denotes the cosine similarity, and $\gamma$ is a scale parameter. Following \cite{ChenLKWH19}, $\gamma=10$ is used. 

\noindent {\bf Meta-Test.} Its workflow is similar to the meta-training phase. The difference is that we remove the meta-optimizer training step defined in Eq.~\ref{eq5_0} and directly perform few-shot classification for novel classes by following Eqs.~\ref{eq5} $\sim$ \ref{eq7}.

\subsection{Meta-Optimizer}
\label{section_3_3}
In the FSL framework described in Section~\ref{section3_2}, the key challenge is how to design a meta-optimizer to polish prototypes. In this subsection, we first discuss the limitation of existing meta-optimizers when using them to polish prototypes. Then, a novel meta-optimizer, \emph{i.e.}, MetaNODE, is presented. 

\noindent {\bf Limitations.} In the above framework, several existing meta-optimizers can be utilized to diminish prototype bias by setting the prototypes as their variables to be updated, such as MetaLSTM \cite{RaviL17} and ALFA \cite{BaikCCKL20}. However, we find that they suffer from a new drawback, \emph{i.e.}, gradient bias issue. Here, we take ALFA as an example to illustrate the issue. Formally, for each few-shot task, let $L(p(t))$ be its differentiable loss function with prototype $p(t)$, and $\nabla L(p(t))$ be its prototype gradient, 
during performing prototype optimization. Then, the ALFA can be expressed as the following $M$-steps iterations, given the initial (\emph{i.e.}, mean-based) prototypes $p(0)$. That is,
\begin{equation}
	p(t+1) = p(t) - \eta (\nabla L(p(t))+\omega p(t)),
	\label{eq1}
\end{equation}
where $t$ is the iteration step (\emph{i.e.}, $t=0, 1, ..., M-1$), $\eta$ is a learning rate, and $\omega$ denotes a weight of $\ell_2$ regularization term infered by the meta-optimizer $g_{\theta_g}()$. Its goal is to improve fast adaptation with few examples by learning task-specific $\ell_2$ regularization term. Though ALFA is effective (see Table~\ref{table5}), we find that it computes the gradient $\nabla L(p(t))$ in an average manner over few labeled samples $(x_i, y_i) \in \mathcal{S}$:
\begin{equation}
	\nabla L(p(t)) = \frac{1}{|\mathcal{S}|} \sum_{(x_i,y_i) \in \mathcal{S}} \nabla L_{(x_i,y_i)}(p(t)),
	\label{eq2}
\end{equation}
where $|\cdot|$ denotes the size of a set and $\nabla L_{(x_i,y_i)}(p(t))$ is the gradient of the sample $(x_i,y_i) \in \mathcal{S}$. Such estimation is inaccurate, because the number of available labeled (support) samples (\emph{e.g.}, $K$=1 or 5) is far less than the expected amount. As a result, the optimization performance of existing methods is limited. This is exactly the gradient bias issue mentioned in the section of introduction. Note that the unlabeled samples $x \in \mathcal{Q}'$ are not used in these methods because their gradients cannot be computed without labels.

\begin{figure}[t]
	\centering
	\includegraphics[width=0.95\columnwidth]{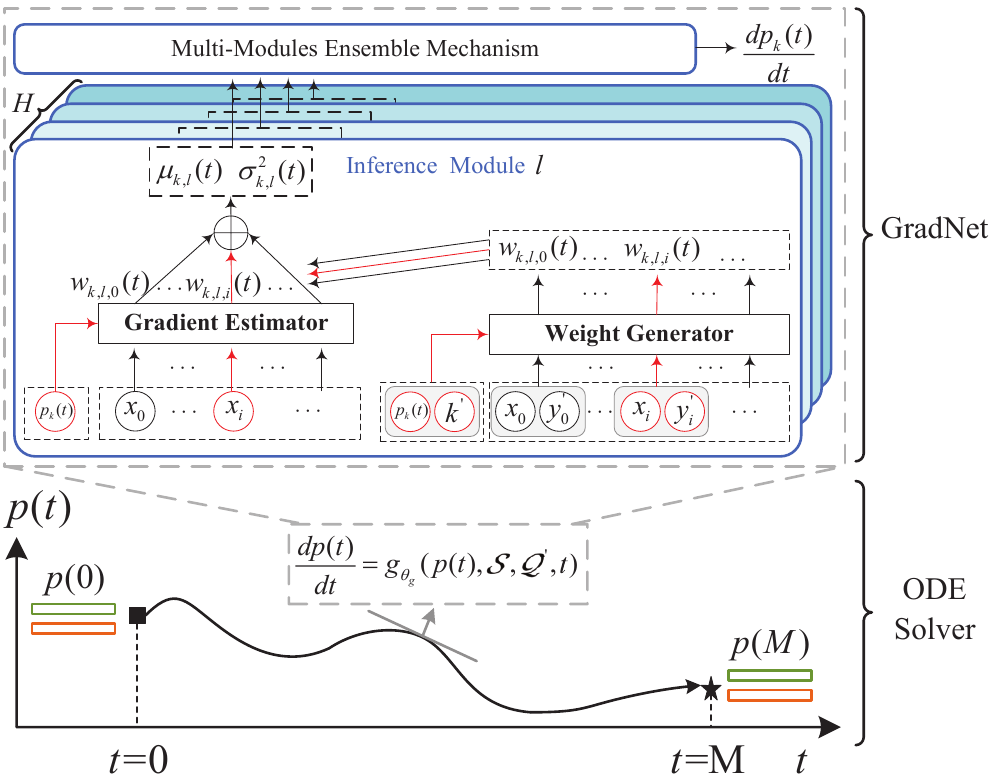} 

	\caption{Illustration of MetaNODE, which characterizes the prototype optimization dynamics by using a Neural ODE.} 
	\vspace{-9.5pt}
	\label{fig3}
\end{figure}

\noindent {\bf MetaNODE.} Recent studies \cite{bu2020} found that the iteration process of Gradient Descent Algorithm (GDA) can be viewed as an Euler discretization of an ordinary differential equation (ODE). That is,
\begin{equation}
	\frac{\mathrm{d} p(t)}{\mathrm{d} t} = - \nabla L(p(t)),
\end{equation}
where $t$ is a continous variable (\emph{i.e.}, time) and $\frac{\mathrm{d} p(t)}{\mathrm{d} t}$ denotes a continuous-time gradient flow of prototypes $p(t)$. 

Inspired by this fact, to more finely rectify the prototypes, we propose to characterize the prototype dynamics by an ODE and consider the prototype rectification problem as the ODE initial value problem, where the initial and final status values correspond to the mean-based and optimal prototypes, respectively. To address the gradient bias issue appeared in Eq. \ref{eq2}, we view the prototype $p(t)$, support set $\mathcal{S}$, unlabeled sample set $\mathcal{Q}'$, and time $t$ as inputs and then employ a neural network (\emph{i.e.} the meta-learner $g_{\theta_g}$()) to directly estimate the continuous gradient flow $\frac{\mathrm{d} p(t)}{\mathrm{d} t}$. Then, the ODE turns into a Neural ODE, \emph{i.e.}, $\frac{\mathrm{d} p(t)}{\mathrm{d} t} = g_{\theta_g}(p(t), \mathcal{S}, \mathcal{Q}', t)$. 

Based on this notion, we design a novel Neural ODE-based Meta-Optimizer (called MetaNODE). Its advantage is that the prototype rectification dynamics can be captured in a continuous manner, thereby more finely diminishing the prototype bias. As shown in Figure~\ref{fig3}, the MetaNODE consists of a Gradient Flow Inference Network (GradNet) and an ODE solver. The GradNet is regarded as the meta-learner $g_{\theta_g}()$, aiming to infer the continuous gradient flow $\frac{\mathrm{d} p(t)}{\mathrm{d} t}$ (see Section~\ref{section3_4} for its details). Based on the GradNet $g_{\theta_g}()$ and initial prototypes $p(0)$, the optimal prototypes $p(M)$ can be obtained by evaluating the Neural ODE at the last time point $t=M$, \emph{i.e.}, $p(M) = p(0) + \int_{t=0}^M g_{\theta_g}(p(t), \mathcal{S}, \mathcal{Q}', t)$, where the integral term is calculated by the ODE solvers. That is,
\begin{equation}
	p(M) = ODESolver(g_{\theta_g}(), p(0), \mathcal{S}, \mathcal{Q}', t=M).
	\label{eq3}
\end{equation}
Following \cite{ChenRBD18}, we use Runge-Kutta method \cite{Alexander90} as our ODE solver because it has relatively low integration error and computational complexity. 

\subsection{Gradient Flow Inference Network}
\label{section3_4}

In this section, we introduce how the GradNet $g_{\theta_g}()$ employed in MetaNODE is designed. Intuitively, different classes have distinct prototype dynamics. For many classes like animals and plants, the differences of their prototype dynamics may be quite large. To model the class diversities, as shown in Figure~\ref{fig3}, instead of performing a single inference module, we design multiple inference modules with the same structure to estimate the prototype gradient $\frac{\mathrm{d} p(t)}{\mathrm{d} t}$. Here, the inference module consists of a gradient estimator and a weight aggregator. The former aims to predict the contributed gradient of each sample $x_i \in \mathcal{S} \cup \mathcal{Q}'$ for prototypes $p(t)$. The latter accounts for evaluating the importance of each sample and then combining their gradient estimations in a weighted mean manner. For clarity, we take inference module $l$ and class $k$ as an example to detail them. 

\noindent {\bf Gradient Estimator.} As we adopt the cosine-based classifier, the prototype is expected to approach the angle center of each class. To eliminate the impact of vector norm, we first transform the features $f_{\theta_f}(x_i)$ of each sample $x_i$ to an appropriate scale by a scale layer $g_{\theta_{gs}^{l}}()$ with parameters $\theta_{gs}^{l}$. Then, the gradient $d_{k, l, i}(t)$ is estimated by computing the difference vector between it and prototype $p_k(t)$. That is,
\begin{equation} 
	\begin{aligned}
		d_{k, l, i}(t) = g_{\theta_{gs}^{l}}(f_{\theta_f}(x_i)\|p_k(t)) \otimes f_{\theta_f}(x_i) - p_k(t),
	\end{aligned}
	\label{eq12}
\end{equation}
where $\|$ is a concatenation operation of two vectors and $\otimes$ denotes an element-wise product operation.

\noindent {\bf Weight Generator.} Intuitively, different samples make varying contributions to the gradient prediction of prototype $p_k(t)$. To this end, we design a weight generator to predict their weights. Specially, for each sample $x_i \in \mathcal{S} \cup \mathcal{Q}'$, we combine the prototype $p_k(t)$ and the sample $x_i$ as a new feature. Then, the weight generating process involves a simple feed-forward mapping of the new features by an embedding layer $g_{\theta_{ge}^{l}}()$, followed by a relation layer $g_{\theta_{gr}^{l}}()$ with a multi-head based attention mechanism \cite{VaswaniSPUJGKP17} and an output layer $g_{\theta_{go}^{l}}()$. Here, the relation layer aims to obtain a robust representation by exploring the pair-wise relationship between all samples and the output layer evaluates the contributed weight $w_{k,l,i}$. The above weight generating process can be summarized as Eq.~\ref{eq13}. That is,
\begin{small}
\begin{equation} 
	\begin{aligned}
		h_{k,l,i}(t)=g_{\theta_{ge}^{l}}(k'\|p_k(t)\|&y'_i\|f_{\theta_f}(x_i)\|p_k(t) \otimes f_{\theta_f}(x_i)),
		\\h_{k,l,i}'(t) =  g_{\theta_{gr}^{l}}&(\{h_{k,l,i}(t)\}_{i=0}^{|\mathcal{S} \cup \mathcal{Q}|-1}),
	    \\w_{k,l,i}(t) = &g_{\theta_{go}^{l}}(h_{k,l,i}'(t)),
	\end{aligned}
	\label{eq13}
\end{equation}
\end{small}where $\theta_{ge}^{l}$, $\theta_{gr}^{l}$, and $\theta_{go}^{l}$ denote model parameters; $k'$ and $y'_i$ denotes the one-hot label of prototype $p_k(t)$ and sample $x_i$, respectively. We replace the one-hot label of each unlabeled sample $x_i \in \mathcal{Q}'$ in a $N$-dim vector with value of $1/N$.

Finally, the gradient $\mu_{k,l}(t)$ and its estimation variance $\sigma^{2}_{k,l}(t)$ can be obtained in a weighted mean manner:
\begin{footnotesize}
	\begin{equation} 
		\begin{aligned}
			\mu_{k,l}(t)=&\sum\nolimits_{i} w_{k,l,i}(t) \otimes d_{k,l,i}(t), \\ \sigma_{k,l}^2(t)=\sum\nolimits_{i} &w_{k,l,i}(t) \otimes (d_{k,l,i}(t) -\ \mu_{k,l}(t))^2.
		\end{aligned}
		\label{eq17}
	\end{equation}
\end{footnotesize}

\noindent {\bf Multi-Modules Ensemble Mechanism.} We have obtained multiple gradient estimations for prototype $p_k(t)$, \emph{i.e.}, $\{\mu_{k,l}(t)\}_{l=0}^{H-1}$, where $H$ denotes the number of inference modules. 
Intuitively, the variance $\{\sigma_{k,l}^{2}(t)\}_{l=0}^{H-1}$ reflects the inconsistency of gradients contributed by all samples, \emph{i.e.}, the larger variance implies greater uncertainty. Hence, to obtain a more reliable prototype gradient $\frac{\mathrm{d} p_k(t)}{\mathrm{d} t}$, we regard the variances as weights to combine these gradients $\mu_{k,l}(t)$: 
\begin{scriptsize}
\begin{equation} 
	\begin{aligned}
		\frac{\mathrm{d} p_k(t)}{\mathrm{d} t}=\beta\left[\sum_{l=0}^{H-1} (\sigma_{k,l}^{2}(t))^{-1}\right]^{-1}\left[\sum_{l=0}^{H-1} (\sigma_{k,l}^{2}(t))^{-1} \mu_{k,l}(t)\right],
	\end{aligned}
	\label{eq19}
\end{equation}
\end{scriptsize}where we employ a term of exponential decay with time $t$, \emph{i.e.}, $\beta = \beta_0 \xi^{\frac{t}{M}}$ ($\beta_0$ and $\xi$ are hyperparameters, which are all set to 0.1 empirically), to improve the model stability. 

\subsection{Adaption to Inductive FSL Setting}
\label{section3_5}

The above MetaNODE-based framework focuses on the transductive FSL, which can also be easily adapted to inductive FSL. Specifically, its workflow is similar to the process described in Sections~\ref{section3_2} and \ref{section_3_3}. The only difference is the unlabeled sample set $\mathcal{Q}'$ is removed in the GradNet of Sections~\ref{section3_4}, and only using support set $\mathcal{S}$ to estimate gradients.

\section{Performance Evaluation}


\begin{table*}
	\begin{center}
		\smallskip\scalebox
		{0.74}{
			\begin{tabular}{l|l|c|c|c|c|c|c}
				\hline
				\multicolumn{1}{l|}{\multirow{2}{*}{Setting}}&\multicolumn{1}{c|}{\multirow{2}{*}{Method}}&\multicolumn{1}{c|}{\multirow{2}{*}{Type}}&\multicolumn{1}{c|}{\multirow{2}{*}{Backbone}}& \multicolumn{2}{c|}{miniImagenet} & \multicolumn{2}{c}{tieredImagenet} \\ 
				\cline{5-8}
				& & & & 5-way 1-shot & 5-way 5-shot & 5-way 1-shot & 5-way 5-shot \\
				\hline \hline
				\multicolumn{1}{l|}{\multirow{10}{*}{\makecell*[c]{Trans\\ductive}}} 
				& DPGN \cite{YangLZZZL20} & Graph & ResNet12 &  $67.77 \pm 0.32\%$  & $84.60 \pm 0.43\%$ & $72.45 \pm 0.51\%$  & $87.24 \pm 0.39\%$ \\
				& EPNet \cite{RodriguezLDL20} & Graph & ResNet12 & $66.50 \pm 0.89\%$  & $81.06 \pm 0.60\%$ & $76.53 \pm 0.87\%$  & $87.32 \pm 0.64\%$ \\
				& MCGN \cite{Tang_2021_CVPR} & Graph & Conv4 & $67.32 \pm 0.43\%$  & $83.03 \pm 0.54\%$ & $71.21 \pm 0.85\%$  & $85.98 \pm 0.98\%$ \\
				& ICI \cite{WangXLZF20} & Pre-training& ResNet12 & $65.77 \%$  & $78.94 \%$ & $80.56 \%$  & $87.93 \%$ \\
				& LaplacianShot \cite{ZikoDGA20} & Pre-training & ResNet18 & $72.11 \pm 0.19\%$  & $82.31 \pm 0.14\%$ & $78.98 \pm 0.21\%$  & $86.39 \pm 0.16\%$ \\
				& SIB \cite{HuMXSOLD20} & Pre-training & WRN-28-10 & $70.0 \pm 0.6\%$  & $79.2 \pm 0.4\%$ & -  & - \\
				& BD-CSPN \cite{YaohuiWang_pr} & Pre-training & WRN-28-10 &  $70.31 \pm 0.93\%$  & $81.89 \pm 0.60\%$ & $78.74 \pm 0.95\%$  & $86.92 \pm 0.63\%$ \\	
				& SRestoreNet \cite{XueW20} & Pre-training & ResNet18 &  $61.14 \pm 0.22\%$  & - & -  & - \\	
				& ProtoComNet \cite{zhang2021} & Pre-training & ResNet12 &  $73.13 \pm 0.85\%$  & $82.06 \pm 0.54\%$ & $81.04 \pm 0.89\%$  & $87.42 \pm 0.57\%$ \\
				\cline{2-8}
				& MetaNODE (ours) & Pre-training & ResNet12 & \emph{\textbf{77.92}} $\pm$ \emph{\textbf{0.99}}$\%$ & \emph{\textbf{85.13}} $\pm$ \emph{\textbf{0.62}}$\%$ & \emph{\textbf{83.46}} $\pm$ \emph{\textbf{0.92}}$\%$ & \emph{\textbf{88.46}} $\pm$ \emph{\textbf{0.57}}$\%$ \\	
				\hline
				\multicolumn{1}{l|}{\multirow{9}{*}{\makecell*[c]{In\\ductive}}} 
				& CTM \cite{li2019finding} & Metric & ResNet18 &  $62.05 \pm 0.55\%$  & $78.63 \pm 0.06\%$ & $64.78 \pm 0.11\%$  & $81.05 \pm 0.52\%$ \\
				& ALFA \cite{BaikCCKL20} & Optimization & ResNet12 & $59.74 \pm 0.49\%$  & $77.96 \pm 0.41\%$ & $64.62 \pm 0.49\%$  & $82.48 \pm 0.38\%$ \\
				& sparse-MAML \cite{Johannes21} & Optimization & Conv4 & $56.39 \pm 0.38\%$  & $73.01 \pm 0.24\%$ & $-$  & $-$ \\
				& Neg-Cosine \cite{LiuCLL0LH20} & Pre-training & ResNet12 &  $63.85 \pm 0.81\%$  & $81.57 \pm 0.56\%$ & -  & - \\
				& P-Transfer \cite{ShenLQSC21} & Pre-training & ResNet12 &  $64.21\pm 0.77\%$  & $80.38 \pm 0.59\%$ & -  & - \\
				& Meta-UAFS \cite{Zhiuncertainty21} & Pre-training & ResNet12 &  $64.22\pm 0.67\%$  & $79.99 \pm 0.49\%$ & $69.13 \pm 0.84\%$  & $84.33 \pm 0.59\%$ \\
				& RestoreNet \cite{XueW20} & Pre-training & ResNet12 &  $59.28 \pm 0.20\%$  & - & -  & - \\
				& ClassifierBaseline \cite{Yinbo20} & Pre-training & ResNet12 & 61.22 $\pm$ 0.84$\%$ & 78.72 $\pm$ 0.60$\%$ & 69.71 $\pm$ 0.88$\%$ & 83.87 $\pm$ 0.64$\%$ \\
				\cline{2-8}	
				& MetaNODE (ours) & Pre-training & ResNet12 & 66.07 $\pm$ 0.79$\%$ & 81.93 $\pm$ 0.55$\%$ & 72.72 $\pm$ 0.90$\%$ & 86.45 $\pm$ 0.62$\%$ \\			
				\hline
		\end{tabular}}
	\end{center}
	\caption{Experiment results on miniImageNet and tieredImageNet. The best results are highlighted in bold.}
	\vspace{-9.5pt}
    \label{table1}
\end{table*} 

\begin{table}
	\begin{center}
		\smallskip\scalebox
		{0.75}{
			\begin{tabular}{c|l|c|c}
				\hline
				\multicolumn{1}{l|}{\multirow{2}{*}{Setting}}&\multicolumn{1}{c|}{\multirow{2}{*}{Method}}& \multicolumn{2}{c}{CUB-200-2011} \\ 
				\cline{3-4}
				& & 5-way 1-shot & 5-way 5-shot \\
				\hline \hline
				\multicolumn{1}{l|}{\multirow{7}{*}{\makecell*[c]{Trans\\ductive}}}
				& EPNet &$82.85 \pm 0.81\%$  &$91.32 \pm 0.41\%$\\
				& ECKPN &$77.43 \pm 0.54\%$  &$92.21 \pm 0.41\%$\\
				& ICI & $87.87\%$  & $92.38\%$ \\
				& LaplacianShot & $80.96\%$  & $88.68\%$ \\
				& BD-CSPN &  $87.45\%$  & $91.74\%$ \\
				& RestoreNet &  $76.85 \pm 0.95\%$  & - \\		
				\cline{2-4}
				& MetaNODE (ours) & \emph{\textbf{90.94}} $\pm$ \emph{\textbf{0.62}}$\%$ & \emph{\textbf{93.18}} $\pm$ \emph{\textbf{0.38}}$\%$ \\		
				\hline
				\multicolumn{1}{l|}{\multirow{5}{*}{\makecell*[c]{In\\ductive}}} &				
				RestoreNet &  $74.32 \pm 0.91\%$  & - \\
				& Neg-Cosine &  $72.66 \pm 0.85\%$  & $89.40 \pm 0.43\%$ \\
				& P-Transfer &  $73.88 \pm 0.87\%$  & $87.81 \pm 0.48\%$ \\
				& ClassifierBaseline &  $74.96 \pm 0.86\%$  & $88.89 \pm 0.43\%$ \\
				\cline{2-4}
				& MetaNODE (ours) & 80.82 $\pm$ 0.75$\%$ & 91.77 $\pm$ 0.49$\%$ \\					
				\hline
		\end{tabular}}
	\end{center}
	\caption{Experiment results on CUB-200-2011.}
	\vspace{-9.5pt}
	\label{table2}
\end{table}

\subsection{Datasets and Settings}
\noindent \textbf{MiniImagenet.} The dataset consists of 100 classes, where each class contains 600 images. Following the standard split in \cite{Yinbo20}, we split the data set into 64, 16, and 20 classes for training, validation, and test, respectively. 

\noindent \textbf{TieredImagenet.}
The dataset is a larger dataset with 608 classes. Each class contains 1200 images. Following \cite{Yinbo20}, the dataset is split into 20, 6, and 8 high-level semantic classes for training, validation, and test, respectively. 

\noindent \textbf{CUB-200-2011.}
The dataset is a fine-grained bird recognition dataset with 200 classes. It contains about 11,788 images. Following the standard split in \cite{ChenLKWH19}, we split the data set into 100 classes, 50 classes, and 50 classes for training, validation, and test, respectively. 

\subsection{Implementation Details}
\noindent \textbf{Network Details.}
We use ResNet12 \cite{Yinbo20} as the feature extractor.
In GradNet, we use four inference modules to estimate the gradient flow. For each module, we use a two-layer MLP with 512-dimensional hidden layer for the scale layer, a single-layer perceptron with 512-dimensional outputs for the the embedding layer, a multi-head attention module with $8$ heads and each head contains 16 units for the relation layer, and a single-layers perceptron and a softmax layer for the output layer. ELU \cite{ClevertUH15} is used as the activation function.

\noindent \textbf{Training details.}
Following \cite{Yinbo20}, we use an SGD optimizer to train the feature extractor for 100 epochs. 
In the meta-training phase, we train the Neural ODE-based meta-optimizer 50 epochs using Adam with a learning rate of 0.0001 and a weight decay of 0.0005, where the learning rate is decayed by 0.1 at epochs 15, 30, and 40, respectively. 

\noindent \textbf{Evaluation.} Following \cite{Yinbo20}, we evaluate our method on 600 randomly sampled episodes (5-way 1/5-shot tasks) from the novel classes and report the mean accuracy together with the 95\% confidence interval. In each episode, we randomly sample 15 images per class as the query set.

\subsection{Experimental Results}
We evaluate our MetaNODE-based prototype optimization framework and various state-of-the-art methods on general and fine-grained few-shot image recognition tasks. Among these methods, ClassifierBaseline, RestoreNet, BD-CSPN, SIB, and SRestoreNet are our strong competitors since they also focus on learning reliable prototypes for FSL.

\noindent {\bf General Few-Shot Image Recognition.} Table~\ref{table1} shows the results of various evaluated methods on miniImagenet and tieredImagenet. In transductive FSL, we found that (\romannumeral1) MetaNODE outperforms our competitors (\emph{e.g.}, BD-CSPN, SIB, SRestoreNet) by around 2\% $\sim$ 7\%. This is because we rectify prototypes in a continuous manner; (\romannumeral2) MetaNODE achieves superior performance over other state-of-the-art methods. Different from these methods, our method focuses on polishing prototypes instead of label propagation or loss evaluation. These results verify the superiority of MetaNODE; (\romannumeral3) the performance improvement is more conspicuous on 1-shot than 5-shot tasks, which is reasonable because the prototype bias is more remarkable on 1-shot tasks. 

In inductive FSL, our MetaNODE method outperforms ClassifierBaseline by a large margin, around 3\% $\sim$ 5\%, on both datasets. This means that our method introducing a Neural ODE to polish prototype is effective. MetaNODE outperforms our competitors (\emph{i.e.}, RestoreNet) by around 7\%, which validates the superiority of our manner to rectify prototypes. Besides, MetaNODE achieves competitive performance over other state-of-the-art methods. Here, (\romannumeral1) different from the metric and optimization methods, our method employs a pre-trained feature extractor and focuses on polishing prototype; (\romannumeral2) compared with other pre-training methods, our method focuses on obtaining reliable prototypes instead of fine-tuning feature extractors for FSL. 

\noindent {\bf Fine-Grained Few-Shot Image Recognition.} The results on CUB-200-2011 are shown in Table 2. Similar to Table 1, we observe that MetaNODE significantly outperforms the state-of-the-art methods, achieving 2\% $\sim$ 6\% higher accuracy scores. This further verifies the effectiveness of MetaNODE in the fine-grained FSL task, which exhibits smaller class differences than the general FSL task. 

\subsection{Statistical Analysis}
\label{section_4_4}
\noindent {\bf Does MetaNODE obtain more accurate prototypes?} In Table~\ref{table3}, we report the cosine similarity between initial (optimal) prototypes, \emph{i.e.}, $p(0)$ ($p(M)$) and real prototypes on 5-way 1-shot tasks of miniImagenet. The real prototypes are obtained by averaging the features of all samples $x_i \in \mathcal{S} \cup \mathcal{Q}$ by following \cite{YaohuiWang_pr}. 
The results show that MetaNODE obtains more accurate prototypes, which is because MetaNODE regards it as an optimization problem, and solves it in a continuous dynamics-based manner.

\noindent {\bf Does MetaNODE alleviate gradient bias?} In Table~\ref{table4}, we randomly select 1000 episodes from miniImageNet, and then calculate the cosine similarity between averaged (inferred) and real gradient. Here, the averaged and infered gradients are obtained by Eq.~\ref{eq2} and GradNet, respectively. The real gradients are obtained in Eq.~\ref{eq2} by using all samples. We select SIB as the baseline, which improves GDA by inferring the loss value of unlabeled samples. It can be observed that MetaNODE obtains more accurate gradients than SIB. This is because we model and train a meta-learner from abundant FSL tasks to directly estimate the continuous gradient flows.

\begin{figure}
	\vspace{-11.5pt}
	\centering
	\subfigure[Performance]{ 
		\label{fig4a} 
		\includegraphics[width=0.48\columnwidth]{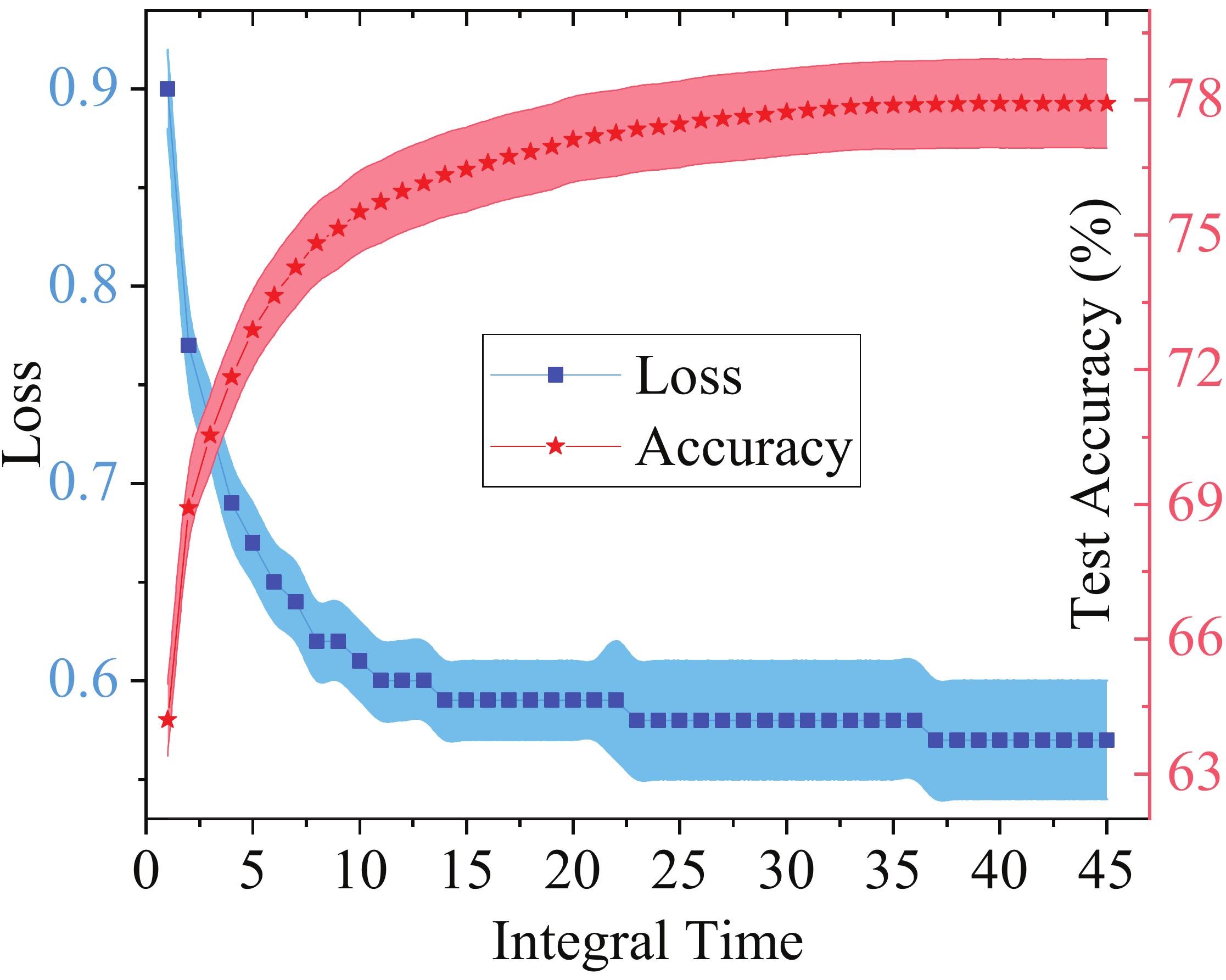}}
	\subfigure[Feature space]{ 
		\label{fig4b} 
		\includegraphics[width=0.48\columnwidth]{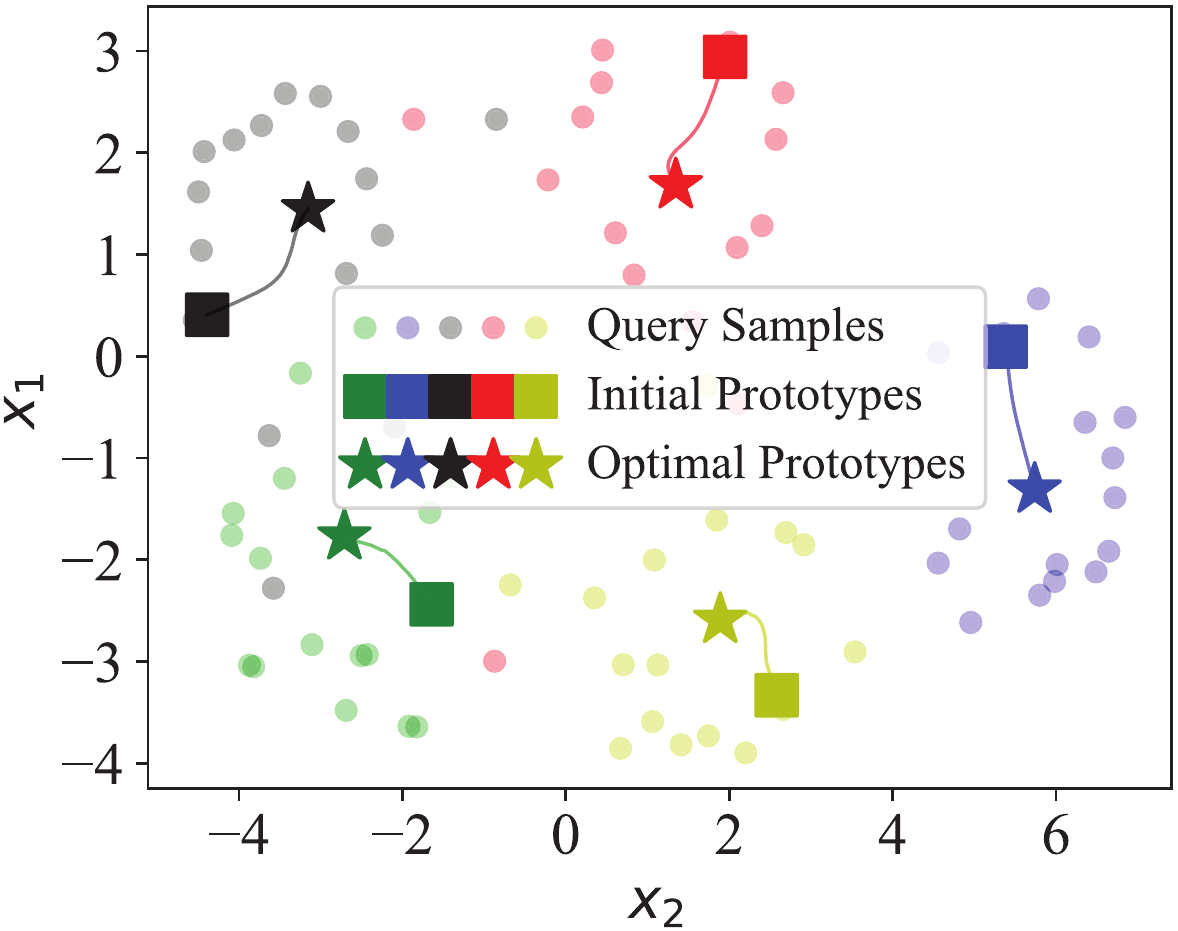}}
	\vspace{-11.4pt}
	\caption{Visualization of Neural ODE on miniImagenet.}
	\label{fig4}
\end{figure}

\noindent {\bf Can our meta-optimizer converge?} In Figure~\ref{fig4a}, we randomly select 1000 episodes (5-way 1-shot) from miniImageNet, and then report their test accuracy and loss from integral time $t = $ 1 to 45. It can be observed that our meta-optimizer can converge to a stable result when $t=40$. Hence, $M=40$ is a default setting in our approach.  

\noindent {\bf How our meta-optimizer works?}
We visualize the prototype dynamics of a 5-way 1-shot task of miniImagenet, in Figure~\ref{fig4b}. Note that (1) since there is only one support sample in each class, its feature is used as the initial prototypes; (2) the curve from the square to the star denotes the trajectory of prototype dynamics and its tangent vector represents the gradient predicted by our GradNet. We find that the initial prototypes marked by squares flow to optimal prototypes marked by stars along prototype dynamics, much closer to the class centers. This indicates that our meta-optimizer effectively learns the prototype dynamics. 

\begin{table}	
	\centering
	\smallskip\scalebox
	{0.80}{
		\smallskip\begin{tabular}{l|c|c}
			\hline
			Methods & Initial Prototypes & Optimal Prototypes \\
			\hline \hline
			BD-CSPN & 0.64 & 0.75 \\
			SRestoreNet & 0.64 & 0.85 \\
			ProtoComNet & 0.64 & 0.91 \\
			\hline
			MetaNODE & 0.64 & 0.93 \\
			\hline
	\end{tabular}}
	\caption{Experiments of prototype bias on miniImagenet. }\smallskip
	\vspace{-11.5pt}
	\label{table3}
\end{table} 

\begin{table}
	\centering
	\smallskip\scalebox
	{0.80}{
		\smallskip\begin{tabular}{l|c|c}
			\hline
			Methods & Averaged Gradient & Infered Gradient \\
			\hline \hline
			SIB & 0.0441 & 0.0761 \\
			MetaNODE & 0.0441 & 0.1701 \\
			\hline
	\end{tabular}}
	\caption{Experiments of gradient bias on miniImagenet. }\smallskip
	\vspace{-11.5pt}
	\label{table4}
\end{table}

\subsection{Ablation Study}
\label{section_4_5}

\noindent {\bf Is our meta-optimizer effective?} In Table~\ref{table5}, we analyzed the effectiveness of our meta-optimizer. Specifically, in inductive FSL setting, (\romannumeral1) we remove our meta-optimizer as a baseline; 
(\romannumeral2) we add the MetaLSTM meta-optimizer \cite{RaviL17} on (\romannumeral1) to optimize prototypes; (\romannumeral3) we replace MetaLSTM by the ALFA \cite{BaikCCKL20} on (\romannumeral2); (\romannumeral4) different from (\romannumeral3), we replace MetaLSTM by our MetaNODE; and (\romannumeral5) we further explore unlabeled samples on (\romannumeral4). From the results of (\romannumeral1) $\sim$ (\romannumeral4), we observe that: 1) the performance of (\romannumeral2) and (\romannumeral3) exceeds (\romannumeral1) around 1\% $\sim$ 3\%, which means that it is helpful to leverage the existing meta-optimizer to polish prototypes and validates the effectiveness of the proposed framework. 2) the performance of (\romannumeral4) exceeds (\romannumeral2) $\sim$ (\romannumeral3) around 2\% $\sim$ 3\%, which shows the superiority of our MetaNODE meta-optimizer. This is because MetaNODE regards the gradient flow as meta-knowledge, instead of hyperparameters like weight decay. Finally, comparing the results of (\romannumeral4) with (\romannumeral5), we find that using unlabeled samples can significantly enhance MetaNODE.

\noindent {\bf Are these key components (\emph{i.e.}, ensemble, attention, and exponential decay) effective in GradNet?} In Table~\ref{table7}, we evaluate the effect of these three components. Specifically, (\romannumeral1) we evaluate MetaNODE on miniImagenet; (\romannumeral2) we remove ensemble mechanism on (\romannumeral1), \emph{i.e.}, $H=1$; (\romannumeral3) we remove attention mechanism on (\romannumeral1); (\romannumeral3) we remove exponential decay mechanism on (\romannumeral1). We find that the performance decreases by around 1\% $\sim$ 3\% when removing these three components, respectively. This result implies that employing these three key components is beneficial for our MetaNODE. 

\begin{table}[t]
	\centering
	\smallskip\scalebox
	{0.75}{
		\smallskip\begin{tabular}{c|l|c|c}
			\hline
			& Method &  5-way 1-shot & 5-way 5-shot \\
			\hline \hline
			(\romannumeral1) & Baseline & 61.22 $\pm$ 0.84$\%$ & $78.72 \pm 0.60\%$  \\
			(\romannumeral2) & + MetaLSTM & 63.85 $\pm$ 0.81$\%$ & 79.49 $\pm$ 0.65$\%$ \\
			(\romannumeral3) & + ALFA & 64.37 $\pm$ 0.79$\%$ & $80.75 \pm 0.57\%$ \\
			(\romannumeral4) & + Neural ODE & 66.07 $\pm$ 0.79$\%$ & 81.93 $\pm$ 0.55$\%$ \\
			(\romannumeral5) & + Neural ODE + QS & 77.92 $\pm$ 0.99$\%$ & $85.13 \pm 0.62\%$ \\
			\hline
	\end{tabular}}
	\caption{Analysis of meta-optimizer on miniImagenet. 
		QS denotes unlabeled (query) samples from query set $Q$. 
	}\smallskip
	\vspace{-10.5pt}
	\label{table5}
\end{table}

\begin{table}[t]
	\centering
	\smallskip\scalebox
	{0.80}{
		\smallskip\begin{tabular}{c|c|c|c}
			\hline
			& Method & 5-way 1-shot & 5-way 5-shot \\
			\hline \hline
			(\romannumeral1) & MetaNODE & 77.92 $\pm$ 0.99$\%$ & 85.13 $\pm$ 0.62$\%$ \\
			(\romannumeral2) & w/o ensemble  & 75.34 $\pm$ 1.10$\%$ & 84.00 $\pm$ 0.53$\%$ \\
			(\romannumeral2) & w/o attention  & 75.10 $\pm$ 0.98$\%$ & 83.90 $\pm$ 0.56$\%$ \\
			(\romannumeral2) & w/o exponential decay  & 76.02 $\pm$ 1.17$\%$ & 84.16 $\pm$ 0.64$\%$ \\
			\hline
	\end{tabular}}
	\caption{Analysis of GradNet components on miniImagenet.}\smallskip
	\vspace{-11.4pt}
	\label{table7}
\end{table}

\section{Conclusion}
In this paper, we propose a novel meta-learning based prototype optimization framework to obtain more accurate prototypes for few-shot learning. In particular, we design a novel Neural ODE-based meta-optimizer to capture the continuous prototype dynamics. Experiments on three datasets show that our model significantly obtains superior performance over state-of-the-art methods. We also conduct extensive statistical experiments and ablation studies, which further verify the effectiveness of our method. 

\section*{Acknowledgments}
This work was supported by the Shenzhen Science and Technology Program under Grant No. JCYJ201805071-83823045, JCYJ20200109113014456, and JCYJ20210324-120208022, and NSFC under Grant No. 61972111. 

\bibliography{egbib}

\end{document}